\documentclass[11pt,a4paper]{article}
\usepackage[hyperref]{acl2020}
\usepackage{arydshln}
\usepackage{times}
\usepackage{latexsym}

\usepackage{microtype}

\aclfinalcopy 

\usepackage{graphicx}
\usepackage{textcomp}
\usepackage[misc]{ifsym}

\title{Exploring aspects of similarity between spoken personal narratives by disentangling them into narrative clause types}

\author{Belen Saldias \\
  MIT Media Lab\\
  Cambridge MA 02139, USA\\
  \texttt{belen@mit.edu} \Letter \\\And
  Deb Roy \\
  MIT Media Lab\\
  Cambridge MA 02139, USA\\
  \texttt{dkroy@media.mit.edu} \\}
\date{}
\begin{document}

\maketitle
\begin{abstract}
Sharing personal narratives is a fundamental aspect of human social behavior as it helps share our life experiences. We can tell stories and rely on our background to understand their context, similarities, and differences. A substantial effort has been made towards developing storytelling machines or inferring characters' features. However, we don't usually find models that compare narratives. This task is remarkably challenging for machines since they, as sometimes we do, lack an understanding of what similarity means. To address this challenge, we first introduce a corpus of real-world spoken personal narratives comprising 10,296 narrative clauses from 594 video transcripts. Second, we ask non-narrative experts to annotate those clauses under Labov's sociolinguistic model of personal narratives (i.e., action, orientation, and evaluation clause types) and train a classifier that reaches 84.7\% F-score for the highest-agreed clauses. Finally, we match stories and explore whether people implicitly rely on Labov's framework to compare narratives. We show that \texttt{actions} followed by the narrator's \texttt{evaluation} of these are the aspects non-experts consider the most. Our approach is intended to help inform machine learning methods aimed at studying or representing personal narratives.

\end{abstract}

\section{Introduction}

We can develop the ability to retrieve a story that we have experienced or heard when someone else is telling a story. We find ourselves thinking about our story, and so we think that we know what is coming next in our friend's story. However, in order for computers to match stories automatically, we need to understand what ``matching'' implies and what aspect of a story should be attended to.

There have been some attempts to match stories \cite{nguyen2014using, chaturvedi2018have} and to understand human judgment about matched stories \cite{nguyen2014using, reagan2016emotional}. Nevertheless, these efforts have been mostly developed in supervised scenarios that already have a set of matched stories in hand, and they are mostly focused on non-personal narratives (e.g., fictional). From these insightful works, however, we want to explore the understanding that when we consider stories to be similar, we attend to some aspects more than others, stressing the need for comparison of different aspects rather than at a global level. 

As a first effort towards our purpose, we collect the largest annotated corpus of spoken personal narratives to our knowledge, comprising 10,296 narrative clauses from 594 stories. We use transcripts of Roadtrip Nation (RTN) videos\footnote{\url{https://roadtripnation.com/}}, where professionals share stories about their lives and career pathways. As for the annotation task, we ask Mechanical Turkers to annotate each clause under Labov's sociolinguistic model of personal narratives \cite{labov1967waletzky}, where a narrative is defined by a structural component, which includes a temporal organization (action clauses) and contextual orientation (orientation clauses), and an evaluation component (evaluation clauses), which represents storytellers'/characters' needs and desires (explained in more depth in section \ref{sec-3}).  

Next, aiming to automatically tag stories, we develop a model to classify these clauses that reaches 84.7\% F-score for the highest-agreed clauses. Once we can automatically differentiate among clause types, we would like to use them to compare stories, but, do ordinary people rely on these clause types to compare narratives? To approach that question, we pair stories and run experiments to understand to what extent ordinary people (as opposed to literary experts) rely on Labov's model to think about similarities among these stories.

Our approach is intended to help inform machine learning methods aimed at studying personal narratives and at modeling abstract information extraction. To the best of our knowledge, this work is the first to propose and develop an approach to understand whether ordinary people rely on Labov's framework to compare personal narratives and what they perceive as similarities among those narratives. We show that actions followed by the narrator’s evaluation of these are the aspects non-experts consider the most when they compare stories. Our main contributions can be summarized as follows:

\begin{itemize}
    \item We acquire annotations to comprehensively label real-world spoken personal narratives, amounting to 10,296 clauses under Labov's clause types, and develop a straightforward strategy to classify those clauses.
    \item We explore to what extent people rely on Labov's framework to compare stories and show that people tend to recognize better similarities in action and evaluation clauses.
    \end{itemize}

The rest of the paper is organized as follows. In section \ref{sec-2}, we present some main related work. In section \ref{sec-3}, we specify the story aspects to be used in our experiments. In section \ref{sec-4}, we describe the uniqueness of our introduced narrative corpus. In section \ref{sec-5} and \ref{sec-6} we describe results, and we end with conclusion and future directions in section \ref{sec-7}.

\section{Related Work}
\label{sec-2}

Our work is preceded by substantial efforts toward document \cite{blei2003latent, dai2015document, yang2016hierarchical} and story \cite{mostafazadeh2016corpus, chaturvedi2018have, iyyer2016feuding, antoniak2019narrative, fu2019asking} representation. We find that most approaches to text similarity focus on non-narrative corpora \cite{vor2019text, lin2013similarity, Cer_2017}. We also observe that most works in stories have been developed for non-personal narratives.

An specific approach to story matching was proposed by \citet{chaturvedi2018have}, who used movie remakes from Wikipedia as paired stories and showed that even in that scenario it was challenging to match the remakes. Additionally, their method does not generalize well to other story types (or even movie plots) since they include specific movie parameters, like characters' name and gender, as the basis of their solution, which does not apply to our case since we do not attempt to match stories based on these surface-level indicators. The closest work to ours was done by \citet{nguyen2014using}, who proposed a set of crowdsourcing tasks to analyze perception of similarity in folk narratives. They tried various approaches to retrieve these narratives. Nevertheless, they had in hand a set of metadata labels that allowed them to match narratives prior to any experiment.

How we narrate our stories was initially studied by \citet{labov1967waletzky}. More recently, \citet{swanson2014identifying} proposed the first mechanism to automatically classify Labov's clauses (action, orientation, and evaluation-type clauses) in personal narratives based on clauses' syntactical structure, namely part-of-speech (POS). By using 50 short stories from online mini-blogs, of diverse topics and structures, they developed a well-defined set of definitions to properly annotate Labov's clause types (referred to as \textit{baseline} method and dataset onward). However, personal narratives from spoken stories set a more challenging context for both annotation and collection (see section \ref{sec-4}). We get inspiration from these works to approach clause types classification using newer techniques like word embeddings \cite{pennington2014glove} and neural networks \cite{kim2014convolutional}.

Furthermore, as we learn to disentangle narrative dimensions or aspects (namely, action, orientation, and evaluation-type clauses), we can use them for other story representation tasks. For instance, identifying the clauses within a story that tell people's intents/desires, reactions, and evaluation of the events (e.g., emotions) can help train and evaluate models aimed at detecting, or planning plots conditioned on, those underlying intentions and reactions \cite{rashkin2018event2mind, guan2019story}.

\section{Story aspects of comparison}
\label{sec-3}

Stories can be thought to be similar in a variety of dimensions; unlike most non-narrative texts, stories have ``meta'' dimensions that go beyond what is said (context of a story, actions that happen, emotional content, speaker's backgrounds, among others). In this work, we explore to what extent Labov's model for personal narratives underlies how non-expert people perceive story similarities. We focus on the following three aspects:

\textit{Temporal organization (action clauses)}: These clauses express a series of events. The narrator might play with the story's chronology, causing differences between narrative structures of one narrator and another.

\textit{Contextual world (orientation clauses)}: These clauses describe information about the context in which actions occur; they serve to orient the audience about people, places, time, and behavioral situations.

\textit{Human needs and desires (evaluation clauses)}: These clauses give significance and tell about the purpose of telling that story; they express the narrator's needs and desires.

\begin{figure}[t!]
    \centering
    \includegraphics[width=\linewidth]{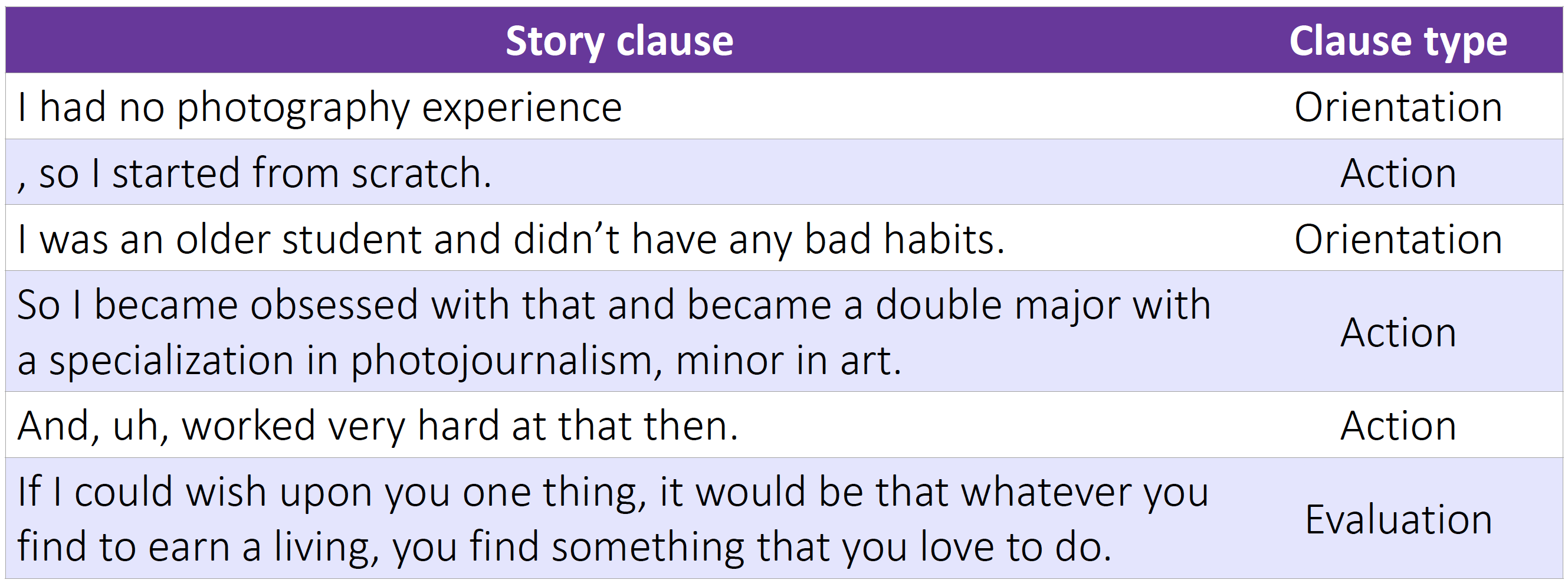}
    \caption{A fragment of a personal narrative in the RTN corpus annotated by Turkers using Labov's model.}
    \label{fig:my_label}
    \vspace{-.2cm}
\end{figure}

See figure \ref{fig:my_label} for an example of a narrative annotated under Labov's model for personal narratives.

\section{Narrative corpus}
\label{sec-4}
We introduce the largest dataset of annotated spoken personal narratives to our knowledge, from now on referenced as Roadtrip Nation or RTN corpus. These narratives were obtained from transcripts of stories video-recorded by Roadtrip Nation (RTN). In those videos, people from many backgrounds share stories about their lives and career pathways. The corpus comprises 10,296 narrative clauses from 594 stories (each one told by a different person), which account for more than 10 hours of people telling stories, each one averaging 17.1 clauses or 62 seconds long, where each clause has on average 11 tokens.

To split narratives into clauses, we proceed as follows. For every sentence in the story, we take every independent clause along with its dependent clause, which account for one narrative clause. To determine clauses, we rely on top-level S* (S, SINV, SBAR, SBARQ, SQ) tags from Penn Treebank II \cite{bies1995bracketing}. For each top-level S* tag, we take its subtree along with hanging prepositions, conjunctions, and adverbs.

While we propose to automatically split our data, \citet{swanson2014identifying}'s data (our \textit{baseline} dataset) was split by trained humans. We compared our strategy implemented using NLTK with their strategy by spliting their stories as well; we found that our method differs at most in one clause from their manually split stories.

{}

\subsection{Uniqueness of this narrative corpus}

This corpus is particularly well-suited to study oral personal narratives for a few reasons. First of all, these stories were all video-recorded and manually transcribed (by Roadtrip Nation (RTN)\footnote{\url{https://roadtripnation.com/}}). These stories are raised from  spontaneous questions during real-world interviews to adults conducted by high school or college students, which produces a fluid and constantly changing dynamic.

Additionally, we recognize the storytellers' awareness of the listeners due to the presence of oral discourse markers that are prominent in oral narratives, such as ``you know,'' ``right,'' ``anyway,'' ``like,'' ``ah,'' ``uh,'' among others. Particularly, ``you know'' is the most frequent bigram in our dataset (0.5\% of all bigrams, 437 appearances) compared to the baseline dataset to study Labov's model, which has ``you know'' mentioned only 3 times throughout all stories \cite{swanson2014identifying}. Furthermore, we find that the word ``you'' appears in the RTN stories an average of 5.3 times per story vs. 1.4 times in the baseline stories.

Besides giving a background (orientation clauses) and telling events (action clauses), RTN stories are specifically produced to display meaningful life experiences or pathway decisions to make the listener reflect or engage with the stories. These purposes emphasize Labov's evaluative function (evaluation clauses) of describing the storyteller's motivation in telling their story.

Here are two randomly sampled full transcripts (i.e., RTN stories), where we can see some of the spectrum of the stories in this corpus:

1. \textit{``In college, I was figuring my life out. I didn't have an exact plan in terms of what I wanted to do. Everybody that acted in the capacity of a guidance counselor to me helped mold me into where I am today. For instance, when I was in high school, my guidance counselor told me , Chris, based on what I know about you, I know you love to be in big cities. I know you love to study human behavior and psychology. We discussed where I might end up in college , so I chose to go to NYU based on that feedback. And when I got my first job in marketing analytics, it's when I realized that hey, this is really cool, I actually really like this. Don't feel like you have to know all the answers right now. The more strict you are in terms of what you think you want to do, the less options you'll have. So think outside the box and keep an open mind.''}

2. \textit{``And slowly and slowly, I started doing small jobs, you know, like, you know, I think one of my first jobs was doing, like, you know, ironing Peter Gabriel's suit and giving him powder for Good Morning America. Like, you know, kind of little things like that. But already, working with musicians , I was like,   `This is is where I belong.'  So, a magical thing happened at this time. I got introduced to Lenny Kravitz , and Lenny Kravitz, at the time, ah, was, uh, a poor, starving musician. Eventually, after working with Lenny for a long time, my work started to grow , and I was working with more and more people and doing other things. So, I realized that ... the next step for me... would be to work on a movie.''}

Note that in written stories (such us the ones in the baseline dataset), all the oral discourse markers present in this last story can be proofread and extracted. However, these are inherent to spontaneous oral narratives.

Finally, even though we ran our experiments prompting Turkers to ``focus on the content and not the speakers' characteristics such as accent or gender'' (first note in full instructions), the released dataset\footnote{\url{https://github.com/social-machines/acl-nuse-personal-narratives}} includes speakers' gender to encourage further analysis across people with different backgrounds but similar stories. From results in section \ref{subsec:mapofaspects}, we estimate that Turkers were rarely biased in their assessment of similarity towards gender because when they were asked to explain why two stories were similar, not one reason related to gender (out of 180 explained reasons).

\subsection{Annotation process}
\label{subsec:annotation-process}

We followed the annotation guidelines, for Labov's model extended label set, constructed by trained researchers in \citet{swanson2014identifying} to explain to Mechanical Turkers how to annotate our clauses. Since both domains of stories (RTN vs. baseline data from \citet{swanson2014identifying}'s) are different, we ran earlier small quality-control experiments to understand whether workers could reach an agreement and, if so, generate labeled stories to add as examples to the task description. Turkers were also invited to provide feedback during these early experiments; after two iterations, we converged to a detailed task description. Finally, each story was assigned to three different workers, and an average of 2.23 workers agreed on every clause.

\begin{table}[b!]
    \centering
    \begin{tabular}{l|cc}
         Clause type & RTN & \textit{baseline}\\\hline
         Action & 26.7\% (2.15) & 24.2\% \\
         Evaluation & 40.0\% (2.29) & 50.0\%\\
         Orientation & 29.7\% (2.24) & 24.2\% \\
         Not story & 3.6\% (2.13) & 1.6\% \\\hline
         Total clauses & 10,296 & 1,602
         
    \end{tabular}
    \caption{Label distribution. Find between parentheses the average agreement for each clause type. Note that the evaluation clause type is the most common clause type in both datasets.}
    \label{tab:label-distribution}
\end{table}{}

Aiming for clean annotations, along with injecting gold examples to reduce randomness, workers were rewarded $\$1.35$ per story, were restricted to living in a English-speaking country, had a HIT Approval Rate $\geq 99$ and Number of HITs Approved $\geq 500$, and had been granted Masters status on the platform. We made annotation tasks full description, some audible stories, and collected data for this task available at \url{https://github.com/social-machines/acl-nuse-personal-narratives}. Gold labels were assigned by simple majority, and for those clauses without agreement, we randomly selected one of the assigned labels by annotators. Find the label distribution in table \ref{tab:label-distribution}.  Overall, we have 9,234 clauses with at least 2 Turkers agreed on them, and 3,495 clauses with 3 Turkers agreed on them.

\section{Narrative clauses classification}
\label{sec-5}

Learning to classify narrative clauses can help us disentangle personal narratives' dimensions. Our specific intention is to understand how this decomposition helps compare stories in different aspects (clause types are assumed to be aspects or dimensions within stories for this work). Additionally, each of these clause types can be used independently for various objectives. For instance, action-type clauses could guide events extraction where, even though the narrator might play with the story's chronology, having these clauses apart can help find causal or temporal orders. Also, identifying orientation-type clause can help create a grounded understanding of the story, where actions and emotions depend on the story's environment described in these clauses. Finally, evaluation-type clauses could bring to surface narrators' mental states, which could push forward research on language models conditioned on mental states \cite{rashkin2018event2mind}.

We propose to use a convolutional neural network (CNN) with max-over-time pooling to classify clauses \citep{zhang2015sensitivity}. More specifically, our model consists of a non-static CNN as in \citet{kim2014convolutional}, where we initialize embeddings using $d=300$-dimensional GloVe pre-trained vectors \cite{pennington2014glove} and concatenate to each vector a one-hot vector (45-dimensional) that encodes POS tags associated with every token. We perform 1-max pooling with ReLU activations over each map generated by filters of sizes 2, 3, and 4; we use 30 filters per size. Then, we use two linear ((90, 45), (45, 3)) layers with dropout of 0.3  before the final softmax layer. We also explored fine-tuning BERT \cite{devlin2018bert} and found that, in most tried scenarios, this simple word-based CNN-based model outperformed BERT in accuracy, maybe due to the small fine-tuning dataset.

We randomly split the RTN dataset into 86\% for training, 7\% for validation, and 7\% for testing, removing the ``not story''-clause type. This gives 7,698 training, 619 test, and 634 validation clauses with agreement $\geq 2$. Our vocabulary has around 6,000 tokens, including an unknown word token we use for uncommon words ($\leq 2$ appearances).

For training, we used 60 epochs and early stopping based on the validation error. We trained with different number of filter, linear layer sizes, batch sizes and learning rates set through experimentation based on performance. We find our best results using Adam with a learning rate of 5e-5 and use batch sizes of 64.

\subsection{Baseline}

We compare our best architecture to the baseline approach proposed by \citet{swanson2014identifying}. To reproduce this baseline, we follow the authors' feature engineering approach and use their data split. By running experiments, we observed correspondence with the top 5 feature-relevance ranking that the baseline model found (POS:IND-VBD being the top 1). This informed our decision of using POS in our proposed approach as well. Note that, originally, the baseline model also included relative clause position within a story (which we are not including here since we mostly care about the clause purpose given its language), lexical semantic categories from LIWC \cite{pennebaker2001linguistic}, dependency relations (DEP), and lexical unigrams (STEM). Using all these engineered features, \citet{swanson2014identifying} reached an F-score of 76.7\% on the cases with the highest annotator agreement. We refrained from using all but part-of-speech (POS) engineered features and still achieved 72.7\% F-score by replicating their approach (see table \ref{tab:f1-scores}).

\subsection{Results}

We report results for models trained and tested with (disjoint) sets composed only of clauses where at least two annotators agreed on their corresponding clause types, and as described in section \ref{subsec:annotation-process}, gold truth labels were assigned by simple majority. 

\begin{table}[b!]
    \centering
    \begin{tabular}{l|cc}
         Model & RTN test & \textit{baseline} test  \\\hline
         CNN - RTN & 84.7\% & $^{*}$62.9\% \\
         SVM - \textit{baseline} & $^{*}$37.1\% & 72.7\% \\
         RF - RTN & 48.3\% & $^{*}$52.5\% \\
         Random (see table \ref{tab:label-distribution}) & 40\% & 50\%
    \end{tabular}
    \caption{F-scores of our model vs. the \textit{baseline} for clauses of highest agreement (= 3)  in test sets. ``- RTN'' (236 clauses) and ``- \textit{baseline}'' (238 clauses) refer to what dataset was used for training and validation. ``$^{*}$'' implies testing in a domain that was not part of the training set (RTN vs. baseline dataset), where we trained in one dataset and predicted on the other. Among the different feature-based models that we tried, a linear l1-penalized support vector machine (SVM) and a random forest (RF) reached highest performance. For clauses with agreement $\geq 2$, we obtained 68.31\% F-score (619 clauses).}
    \label{tab:f1-scores}
    \vspace{-.1cm}
\end{table}{}

Results are shown in table \ref{tab:f1-scores}. Our results demonstrate that a simple CNN with pre-trained embedding and no feature engineering reaches high performance in our RTN dataset. Furthermore, we can see that our proposed model (trained on RTN data) still achieved high performance while evaluated on the baseline test set, even though these datasets are from different domains. On the other hand, the baseline support vector machine (SVM, linear and l1-penalized) \cite{cortes1995support} model performs poorly when evaluated in RTN data, likely because it only uses POS (syntactic) features to represent clauses, and both written (baseline) and spoken (RTN) clauses pose different challenges in syntactical structure. We address these challenges by taking advantage of word embeddings' representational power. \textit{From this, we see that our approach (model and dataset) can be generalized to the baseline dataset better than the other way around}.

Additionally, note that if a model always predicts the most common label (or randomly assigns them), the micro-F1-score (i.e., accuracy) for RTN would be 40\% and for the baseline 50\%. We found that when we used the feature-engineering approach proposed by \citet{swanson2014identifying} in the RTN corpus, the best trained and tested standard model, a random forest with 100 estimators (RF) \cite{breiman2001random}, does not perform well in this new corpus. Though, from table \ref{tab:f1-scores}, we also see that it still does better than random (third vs. fourth row). This result suggests that sentence structure and part-of-speech (POS) do not generalize well to classify narrative clause types, as one would expect from POS being predominant in the top 10 most relevant features in this feature-engineering (original and baseline) approach. While the baseline model found POS features to be highly relevant, since our model uses word embeddings, POS information only contributed 2 -- 3\% to the F-score. Furthermore, these results stress the difference between both story domains: video-recorded spoken narratives (RTN) vs. mini-blog written stories (baseline from \citet{swanson2014identifying}).

To sum up, the fact that a simple CNN performs well on this classification task, as illustrated in table \ref{tab:f1-scores}, reflects the high disentangling power that Labov's model proposes for analyzing spoken personal narratives. Finally, since we can automatically annotate and thus disentangle narrative clauses under this framework, our approach shows to be plausible, so we now proceed to explore aspects of similarity.


\section{People's perception of similarity}
\label{sec-6}
Aiming to understand the aspects (i.e., clause types) that ordinary people attend to the most when they think about similarities among stories, we proceeded as follows. We represent each story as a set of narrative clauses, where each clause is initially encoded into a high-dimensional vector by using the Universal Sentence Encoder (USE) introduced by \citet{cer2018universal}. Next, given stories \textit{s} and \textit{s'}, for each clause in \textit{s} we find the closest clause in cosine similarity in \textit{s'} (\textit{s} $\,\to\,$ \textit{s'}), and vice versa (\textit{s'} $\,\to\,$ \textit{s}), and obtain an average similarity score. Using this mechanism, we match stories only at clause-type subsets (action, evaluation, or orientation-type only). Finally, we sample 60 story pairs with average cosine similarity $\geq 0.5$ for one of the clause types matches. See appendix~\ref{sec:appendix-matched-stories} for some sample matched stories.

For our experiments, we use these 60 stories, which are presented to Turkers in audio form only (as opposed to transcript text). While reading and listening might require different attention spans, since Labov's sociolinguistic model focuses on stories that are produced orally (just like these) and these are short stories -- 62 seconds long on average -- we rely on Turkers' auditory cognitive processing.

\subsection{Annotation task: matching stories}
\label{subsec:matchinng-stories}

We prompted: \textit{``Which one of the following stories, A or B, was the most similar to the main story (and why)?''}. Each main story was annotated twice, switching order for A and B; one of these stories is matched at only one clause type level and the other is randomly selected. Table \ref{tab:exp-sim-1} shows these results.

\begin{table}[h!]
    \centering
    \begin{tabular}{l|c}
         Match only at & \% of times detected\\\hline
         Action & 67.8\% \\
         Evaluation & 60.9\% \\
         Orientation & 48.0\% \\
    \end{tabular}
    \caption{What aspects are paid attention. For those stories matched at the action-clause level, 67.8\% of times Turkers recognized the matched story accurately, and selected the random story the remaining 32.2\% of the times (these action-level matched stories were more than two times recognized correctly than incorrectly). Stories matched in evaluation-type clauses were also recognized accurately 60.9\% of the times, which is 50\% more than those stories that were wrongly recognized (60.9\% vs. 39.1\%). As for orientation-level matches, these were recognized somewhat randomly (48\% of the times Turkers selected the matched stories and 52\% of the times they selected a random story). Some reasons behind mismatches could be (1) that Turkers might be paying attention to other not covered aspects (further explored in section \ref{subsec:mapofaspects}), (2) some randomness on annotations, and (3) the matching strategy.}
    \label{tab:exp-sim-1}
\end{table}{}

From this experiment, we conclude that action and evaluation-type clauses were relevant for non-experts when they compared stories for similarity. Hence, our hypothesis on whether ordinary people rely on these Labov's aspects to compare narratives proved to be true for both action and evaluation aspects of a story but not for the orientation aspect.

\subsection{Map of crowdsourced aspects to Labov's}
\label{subsec:mapofaspects}

Trying to understand how Turkers perceived the different aspects, and where mismatches could possibly come from, for the same 60 stories, we selected the story C that has score $\geq 0.5$ at a given match and has the smallest matching score for the other clause types. We asked \textit{``Explain in what aspects (at least three) are the following personal narratives similar''}, hoping that Turkers would give reasons related to the matched dimensions. Note that with this open-ended question, Turkers were invited to think about \textit{any aspects} that came to mind, thus we did not impose aspects on them beforehand.

Next, we map their responses to Labov's aspects; for example, the explanation \textit{``They both have pessimistic thoughts...''} refers to how a narrator feels or perceives the situation $\,\to \,$ \texttt{evaluation}  clause type. Some results of this mapping strategy are illustrated in table \ref{tab:mapping-results}, and results for this mapping process are summarized in table \ref{tab:exp-sim-2}.

\begin{table}[h!]
    \centering
    \begin{tabular}{l|ccc}
Explanation & Mapped aspects \\\hline

    \begin{tabular}{@{}l@{}}
    \footnotesize{``Both started out in one direction}\\ 
    \footnotesize{and switched to a different field.''}
    \end{tabular} & 
    \footnotesize{\texttt{action}}
    \\\hdashline
    
    \begin{tabular}{@{}l@{}}
    \footnotesize{``Both people spoke about intense}\\ 
    \footnotesize{passion for something.''}
    \end{tabular} & 
    \footnotesize{\texttt{evaluation}}
    \\\hdashline
    
    \footnotesize{``Both were from small towns.''} & 
    \footnotesize{\texttt{orientation}}
    \\\hdashline
    
    \begin{tabular}{@{}l@{}}
    \footnotesize{``Both speakers suggest [ac] \textit{pursuing}}\\
    \footnotesize{\textit{their career goals} [ev] \textit{makes them}} \\
    \footnotesize{\textit{a better person} in real life too.''}
    \end{tabular} & 
    
    \begin{tabular}{@{}c@{}}
    \footnotesize{\texttt{action, }}\\ 
    \footnotesize{\texttt{evaluation}}
    \end{tabular}
    \\\hline
    
    \end{tabular}
    \caption{Examples of mapped explanations. We analyzed every open-ended explanation given by Turkers and mapped them to Labov's model according to what aspects these explanations were mostly referring to. Note that not all explanations were granular, hence, for some of them we highlighted more than one aspect (see fourth row in this table).}
    \label{tab:mapping-results}
\end{table}{}

We show that for action- and evaluation-type clauses, Turkers mentioned aspect of similarity related to these clauses at least twice as often (as relevant) as the less relevant aspect in the matched stories, which (again) proves that these Labov's clause types can work as aspects of similarity.

As for orientation-type clauses, while they are still identified as reasons for similarity as illustrated in table \ref{tab:exp-sim-2}, these are not the main reason to match two stories. We argue that this is due to the nature of our prompts to Turkers, which specifically asked for ``stories'' (section \ref{subsec:matchinng-stories}) or ``narratives'' (section \ref{subsec:mapofaspects}); in ordinary people's mind (i.e., non-narrative experts), both of these concepts might not relate to the physical space or context where events and emotions/intentions happen, causing Turkers to not pay as much attention to them. It might also be that since all RTN stories are within the pathways/inspiration/career domain, people get engaged with that part as opposed to if our domain were more diverse in topics, which would then have led people to recur to the orientation aspect  (background/set-up/place) to match them in the absence of common feelings or similar actions/decisions among stories.

\begin{table}[h!]
    \centering
    \begin{tabular}{l|ccc}
Match at & Action & Evaluation & Orientation \\\hline
Action & \textbf{100}\% & 88\% & 44\% \\
Evaluation & 95\% & \textbf{90}\% & 45\% \\
Orientation & 92\% & 96\% & \textbf{58}\%
    \end{tabular}
    \caption{Aspects referenced in 180 explanations of similarity (3 for each of 60 stories). As expected from results in table \ref{tab:exp-sim-1}, explanations related to action and evaluation aspects are highly present in detected reasons for similarity. We see that, for most story pairs, Turkers gave explanations regarding actions that happen within stories. In particular, for pairs matched at action-clause level, every pair was said to be similar due to similar actions. For evaluation-clause level matches, we find explanations mapped to that aspect twice as often as for the least present aspect (90\% vs. 45\%). Finally, while orientation-type clauses were not perceived as a main similarity aspect (see table \ref{tab:exp-sim-1}), we find that for stories matched at orientation clauses, Turkers recognized this aspect to be a reason for similarity more often than for any other matches (58\% / 45\% = 1.28 times).}
    \label{tab:exp-sim-2}
\end{table}{}

\section{Conclusion}
\label{sec-7}

We introduce the largest corpus of annotated spoken personal narratives, to our knowledge, and develop a straightforward method to classify these narratives' clauses using Labov's sociolinguistic model. Our model's high performance in classification reflects the disentangling power that Labov's model offers for analyzing oral personal narratives. Only by being trained in our introduced corpus, our model performs well in an earlier proposed dataset of written stories. Furthermore, we propose the first attempt to understand whether ordinary people (i.e., non-narrative experts), such as Mechanical Turkers, rely on Labov's model to compare personal stories, and show that these people do rely on two out of three Labov's aspects of narrative. Namely, action-type and evaluation-type clauses are perceived as central aspects of comparison, but the same does not apply to, and remains unresolved for, orientation-type clauses. One natural next step would entail shedding light on how different questions' wording and emphasis, aimed at matching stories, affect what people think of as similarity aspects. We hope that these precursory findings about the aspects that proved to underlie story-matching could also be used in a broader set of tasks, such as finding causal or temporal relationships between events, inferring mental states, or grounding actions and emotions in a story's set-up.

Finally, we acknowledge that we have only scratched the surface of this wonderfully rich space of personal narrative representations and of what people focus on when they compare stories. Our overarching goal, of modeling human judgment of narrative similarity and building a machine capable of replicating that behavior, leaves untouched several questions that future research should explore. For example, what other aspects should be examined to represent personal narratives, how to decide the relative relevance of these aspects, and how to model similarity judgments within aspects.

\section*{Acknowledgments}

The authors would like to thank Roadtrip Nation for collecting and sharing their data. We also thank \citet{swanson2014identifying} who shared their corpus with us, and the anonymous reviewers who provided valuable feedback. We were inspired by researchers at the Laboratory for Social Machines (LSM) at MIT who are passionate about storytelling. This project was funded by LSM Member companies McKinsey \& Company and Twitter.

\bibliography{acl2020}
\bibliographystyle{acl_natbib}

\appendix
\section{Sample matched stories}
\label{sec:appendix-matched-stories}

These stories were matched in the action-clause level (stories A and B, with a similarity score of 0.58), and in the orientation-clause level (stories B and C, score of 0.50). Note that some clauses are not displayed due to space limitations.

\captionsetup{belowskip=2pt,aboveskip=2pt}
\begin{table}[t!]
    \centering
    \begin{tabular}{l|c}
    Narrative clause & Clause type \\\hline

    \begin{tabular}{@{}l@{}}
    \footnotesize{``Did I have the pathway figured out,} \\
    \footnotesize{by no means, no at that time, right?}
    \end{tabular} &
    \footnotesize{\texttt{evaluation}}
    \\\hdashline

    \begin{tabular}{@{}l@{}}
    \footnotesize{So I also got involved in a atmospheric}\\ 
    \footnotesize{chemistry lab}
    \end{tabular} & 
    \footnotesize{\textbf{\texttt{action}}}
    \\\hdashline
    
    \footnotesize{, so nothing to do with animals} & 
    \footnotesize{\texttt{orientation}}
    \\\hdashline
    
    \footnotesize{, but a lot to do with the environment.} & 
    \footnotesize{\texttt{orientation}}
    \\\hdashline
    
    \footnotesize{I loved that, but I was like, well} & 
    \footnotesize{\texttt{evaluation}}
    \\\hdashline
    
    \footnotesize{, I really wanna still apply this to animals.} & 
    \footnotesize{\texttt{evaluation}}
    \\\hdashline
    
    \footnotesize{So I went on to graduate school} & 
    \footnotesize{\textbf{\texttt{action}}}
    \\\hdashline

    \begin{tabular}{@{}l@{}}
    \footnotesize{, and I enjoyed teaching, cuz I also }\\
    \footnotesize{worked as a teaching assistant at CSU} \\
    \footnotesize{Long Beach.''}
    \end{tabular} &
    \footnotesize{\textbf{\texttt{action}}}
    \\\hdashline
    
    \end{tabular}
    \caption*{Story A}

\end{table}{}

\begin{table}[h!]
    \vspace{-.2cm}
    \centering
    \begin{tabular}{l|c}
    Narrative clause & Clause type \\\hline

    \begin{tabular}{@{}l@{}}
    \footnotesize{``When I was in school, I wanted to }\\ 
    \footnotesize{be a doctor.}
    \end{tabular} & 
    \footnotesize{\textbf{\texttt{orientation}}}
    \\\hdashline

    \footnotesize{I went to college} & 
    \footnotesize{\textbf{\texttt{action}}}
    \\\hdashline

    \begin{tabular}{@{}l@{}}
    \footnotesize{and I realized I actually didn't wanna}\\ 
    \footnotesize{be a doctor.}
    \end{tabular} & 
    \footnotesize{\texttt{evaluation}}
    \\\hdashline
    
    \begin{tabular}{@{}l@{}}
    \footnotesize{I wanted to do something more in}\\ 
    \footnotesize{public health.}
    \end{tabular} & 
    \footnotesize{\textbf{\texttt{orientation}}}
    \\\hdashline
    
    \footnotesize{And so I went to graduate school} & 
    \footnotesize{\textbf{\texttt{action}}}
    \\\hdashline
    
    \begin{tabular}{@{}l@{}}
    \footnotesize{and I ultimately got a PhD in } \\
    \footnotesize{international relations and global health} \\
    \footnotesize{cuz I'm interested in this question on} \\
    \footnotesize{sort of a global level.}
    \end{tabular} &
    \footnotesize{\textbf{\texttt{action}}}
    \\\hdashline

    \begin{tabular}{@{}l@{}}
    \footnotesize{So although I started off wanting to be }  \\
    \footnotesize{a doctor and although I never became a} \\
    \footnotesize{doctor, except that I guess I do get to} \\
    \footnotesize{be called Dr. Clinton because I have} \\
    \footnotesize{ a doctorate degree.}
    \end{tabular} &
    \footnotesize{\textbf{\texttt{orientation}}}
    \\\hdashline

    \begin{tabular}{@{}l@{}}
    \footnotesize{I've figured out what my passion is and }  \\
    \footnotesize{how to do that in a way that feels right} \\
    \footnotesize{ for me.''} \\
    \end{tabular} &
    \footnotesize{\texttt{evaluation}}
    \\\hdashline
    
    \end{tabular}
    \caption*{Story B}

\end{table}{}

\begin{table}[h!]
    \centering
    \vspace{-.2cm}
    \begin{tabular}{l|c}
    Narrative clause & Clause type \\\hline

    \footnotesize{``Up until the time I got to college} & 
    \footnotesize{\textbf{\texttt{orientation}}}
    \\\hdashline

    \begin{tabular}{@{}l@{}}
    \footnotesize{, I still had the aspiration to maybe}\\ 
    \footnotesize{ go to medical school.}
    \end{tabular} & 
    \footnotesize{\textbf{\texttt{orientation}}}
    \\\hdashline
    
    \begin{tabular}{@{}l@{}}
    \footnotesize{Until I started to really reality}\\ 
    \footnotesize{hit in that my family wasn't very}\\
    \footnotesize{financially well off}
    \end{tabular} & 
    \footnotesize{\texttt{evaluation}}
    \\\hdashline

    \begin{tabular}{@{}l@{}}
    \footnotesize{, and the reality of the fact that}\\ 
    \footnotesize{medical school costs a lot of money}\\
    \footnotesize{, takes a long time.}
    \end{tabular} & 
    \footnotesize{\texttt{evaluation}}
    \\\hdashline

    \begin{tabular}{@{}l@{}}
    \footnotesize{And then it kinda broad my horizons}\\ 
    \footnotesize{a little bit in that I could explore}\\
    \footnotesize{some other options.''}
    \end{tabular} & 
    \footnotesize{\texttt{action}}
    \\\hdashline
  
    \end{tabular}
    \caption*{Story C}

\end{table}{}

\end{document}